\documentclass[letterpaper]{article} 
\usepackage{aaai2026}

\usepackage[T1]{fontenc}   
\usepackage{textcomp}      

\usepackage{times}  
\usepackage{helvet}  
\usepackage{courier}  
\usepackage[hyphens]{url}  
\usepackage{graphicx} 
\usepackage{natbib}  
\usepackage{caption} 
\usepackage{algorithm}
\usepackage{algorithmic}
\usepackage{booktabs}
\usepackage{amsmath}
\usepackage{amsthm}
\usepackage{amsfonts}
\usepackage{amssymb}
\usepackage{booktabs}
\usepackage{multirow}
\usepackage{subcaption}

\newcommand{\ve}{\boldsymbol}

\usepackage{makecell}

\usepackage[absolute,overlay]{textpos}

\usepackage{newfloat}
\usepackage{listings}
\DeclareCaptionStyle{ruled}{labelfont=normalfont,labelsep=colon,strut=off} 
\floatstyle{ruled}
\newfloat{listing}{tb}{lst}{}
\floatname{listing}{Listing}
\nocopyright 

\title{ACE: Adapting sampling for Counterfactual Explanations}
\author{
    Margarita A. Guerrero and Cristian R. Rojas \thanks{This work was partially supported by the Wallenberg AI, Autonomous Systems and Software Program (WASP) funded by the Knut and Alice Wallenberg Foundation. The authors are with the Division of Decision and Control Systems, KTH Royal Institute of Technology, 100 44 Stockholm, Sweden (e-mails: mags3@kth.se, blak@kth.se, crro@kth.se).}\\
}
\usepackage{bibentry}

\begin{document}
\makeatletter
\let\orig@fnsymbol\@fnsymbol
\def\@fnsymbol#1{}
\maketitle
\let\@fnsymbol\orig@fnsymbol
\makeatother

\maketitle

\begin{abstract}
Counterfactual Explanations (CFEs) interpret machine learning models by identifying the smallest change to input features needed to change the model’s prediction to a desired output. For classification tasks, CFEs determine how close a given sample is to the decision boundary of a trained classifier. Existing methods are often sample-inefficient, requiring numerous evaluations of a black-box model—an approach that is both costly and impractical when access to the model is limited. We propose Adaptive sampling for Counterfactual Explanations (ACE), a sample-efficient algorithm combining Bayesian estimation and stochastic optimization to approximate the decision boundary with fewer queries. By prioritizing informative points, ACE minimizes evaluations while generating accurate and feasible CFEs. Extensive empirical results show that ACE achieves superior evaluation efficiency compared to state-of-the-art methods, while maintaining effectiveness in identifying minimal and actionable changes.\end{abstract}

\section{Introduction}\label{sec:Intro}
Today, Artificial Intelligence (AI) has become an integral part of our lives, impacting both personal and professional decisions. A significant challenge arises when determining whether to trust these increasingly complex machine learning models. To comply with the General Data Protection Regulation (GDPR) \cite{voigt_17}—particularly Article 22 concerning automated decision-making—and other AI-specific data protection laws, organizations must explain how data is being used \cite{eu_19}, and, even in the absence of specific laws, numerous recommendations and guidelines advocate for transparency and explainability in AI \cite{Molnar_22,gilpin_18}.
This need has driven the development of Explainable AI (XAI) approaches \cite{samek_19}, which aim to make AI systems more transparent, trustworthy, and less biased, ensuring that their decisions can be understood and justified.

Questions such as ``Why did the model reject my loan application?" or ``What factors led to a specific diagnosis?" can be answered with Model-Agnostic methods, a subfield of XAI that provides explanations for black-box model outputs. Examples include feature importance methods, such as LIME~\cite{ribeiro_16} and SHAP~\cite{lundberg_17}, which identify the attributes that play a major role in the classifier prediction, and methods like Anchor explanations~\cite{ribeiro_18}, which use if-then rules to highlight conditions under which predictions remain consistent or “anchor” the prediction enough.

When addressing these questions, it is essential to identify the minimal input changes needed to achieve a different prediction. However, the methods discussed above fall short, as they do not directly handle this goal, limiting the actionability of their explanations~\cite{Amir_20}. They also overlook feature correlations, often producing unrealistic data points. To address these issues, Counterfactual Explanations (CFEs)~\cite{wachter_17,verma_24} have emerged as a more actionable alternative. CFEs seek the smallest feasible changes that alter a model’s prediction to a desired outcome. This has inspired a variety of methods targeting simplicity~\cite{sadiku_25}, fairness~\cite{fragkathoulas_24}, and diversity~\cite{mothilal_20}, supported by open-source frameworks like CARLA~\cite{carla_21} and OmniXAI~\cite{yang_22}.

In the context of binary classification, CFEs require only the ability to query the black-box model and observe its binary output, $h$. Since they do not need access to the model internals, CFEs are particularly suitable for scenarios where the model is proprietary or inaccessible, as is often the case when the models are owned by third parties. For instance, consider a bank using a third-party model to determine loan approvals. The model parameters are unavailable, and the bank can only query the model and receive a binary output ($h = 1$ for approval, $h = 0$ for rejection). These interactions are often limited in number or incur a significant cost, as the third party owning the black-box model may charge for each query. In this context, generating actionable counterfactuals enables loan applicants to identify feasible changes (e.g., a salary increase or improving their credit score) to improve their chances of approval with a small number of queries.

Current CFE methods explore the input space via geometric expansions~\cite{laugel_18}, multi-objective evolutionary algorithms~\cite{moc_20, mcd_24, Deb_02}, or constraint-guided search~\cite{mace_20, face_20, lucic_22, chvae_20}. Surrogate-based approaches such as BayCon~\cite{baycon_22} and EI-CFX~\cite{spooner_21} leverage Gaussian Processes~\cite{rasmussen_06} or tree models to guide sample selection. Other recent directions include causal inference~\cite{mahajan_19, carma_24}, program synthesis~\cite{Toni_23}, and amortized strategies for sequential recourse~\cite{verma_22}, broadening the field beyond traditional optimization.

Nevertheless, these methods often neglect the cost of querying $h$, resulting in inefficiencies in scenarios where model evaluations are limited or expensive. Additionally, most approaches rely on large datasets for effective calibration, which may not be practical in real-world applications.

To address the limitations of existing methods, we propose the Adaptive sampling for Counterfactual Explanations (ACE) algorithm for classification tasks. ACE uses Gaussian processes to construct surrogate models and leverages Bayesian optimization~\cite{mockus_89,jones_98} to reduce query counts. It avoids complex hyperparameter tuning via a Gaussian prior and ensures global convergence using the penalty method~\cite{gardner_14,Picheny_16}. ACE further incorporates Monte Carlo sampling~\cite{wilson_18} to prioritize informative points based on their relation to the acquired data. A key strength of ACE is its ability to handle both continuous and categorical features, optimizing a cost function via a hybrid approach: Quasi-Newton methods~\cite{broyden_72} for continuous variables and Branch-and-Bound~\cite{land_60} for discrete ones.

Finally, as demonstrated via extensive benchmark simulations, ACE can work in both low- and high-dimensional spaces, identifying CFEs for datasets with as few as 21 features and as many as 784 features, demonstrating its scalability and robustness across different problems. 

\noindent\textit{To the best of our knowledge, this is the first Bayesian estimation-based algorithm explicitly designed for sample-efficient counterfactual generation, achieving high-quality explanations with significantly fewer model evaluations.}

In summary, the main contributions of this work are:
%
\begin{enumerate} \itemsep -2pt
    \item Proposing ACE, a new and efficient method for generating counterfactual explanations that requires only limited access to the black-box model—via query responses—and yet produces plausible, feasible, and actionable explanations.
    \item Conducting a comprehensive quantitative and qualitative evaluation of ACE, comparing it against state-of-the-art counterfactual explanation methods across multiple real-world datasets and illustrative scenarios.
\end{enumerate}
The remainder of this paper is organized as follows: Section~2 formulates the CFE problem; Section~3 presents the ACE algorithm. Sections~4 and 5 report quantitative and qualitative results, respectively, and Section~6 concludes with final remarks.

\textit{Notation}: Vectors and matrices are written in bold. $[n]$ represents the set of indices from 1 to $n$ and $[a]_+ := \text{max}\{ a, 0\}$. $\boldsymbol{x}_{1:n}$ represents a set of observed data points $\{x_1, \dots, x_n\}$, where each $x_i$ is a point in the input space $\mathcal{X}$. We denote by $C^k$ the set of functions that are $k$ times continuously differentiable.
\section{Problem Statement}
Consider a trained classifier $h \colon \mathcal{X} \to \{ 0, 1 \}$, where $\mathcal{X}$ is an input space endowed with a metric $d$, and let $\tilde{x} \in \mathcal{X}$ be a fixed input, referred to as the \emph{instance}, whose predicted value $\tilde{y} = h(\tilde{x})$ we aim to ``explain''. In this context, we define a \emph{counterfactual explanation} (CFE) as the solution to the following optimization problem:
\begin{equation} \label{eq:optimization}
\underset{x}{\text{minimize}} \quad d(x, \tilde{x}) \quad
     \text{s.t.} \quad x \in \mathcal{S}_{db},
\end{equation}
where $\mathcal{S}_{db} = \{ x \in \mathcal{X}\colon h(x) \neq h(\tilde{x}) \}$ represents the decision boundary of the classifier. The solution of \eqref{eq:optimization} corresponds to identifying the point $x$ closest to $\tilde{x}$ that lies on the decision boundary of the classifier ($\mathcal{S}_{db}$), representing the smallest change needed in the input features to induce a classification flip (i.e., $h(x) \neq h(\tilde{x})$).
\section{ACE Algorithm}
In this section we present our novel algorithm, Adaptive sampling for Counterfactual Explanations (ACE), which relies on Bayesian optimization. To construct ACE, we address two central challenges: formulating a cost function $J$ that captures proximity and structural constraints within the input space $\mathcal{X}$, and approximating the behavior of the black-box classifier $h$ through a continuous surrogate model.

Following standard Bayesian classification~\cite{rasmussen_06}, we assume the binary classifier is a thresholded version of a smooth latent function:
\[
h(x) = \mathbb{H}(f(x)),
\]
where $\mathbb{H}\colon \mathbb{R} \to \{0, 1\}$ is the Heaviside step function centered at $0.5$ (i.e., $\mathbb{H}(a) = 1$ if $a \geq 0.5$, and $\mathbb{H}(a) = 0$ otherwise), and $f \colon \mathcal{X} \to \mathbb{R}$ is a smooth function. We refer to $f$ as the \textit{black-box target function}, which ACE models using Gaussian Processes. This decomposition enables a continuous relaxation of the original problem and allows for the use of gradient-based and surrogate-assisted optimization over mixed input spaces.

The next subsections detail the main components of ACE:
\begin{itemize} \itemsep -1pt
\item reformulating the constrained counterfactual problem via a penalized objective over $f$;
\item modeling $f$ with a Gaussian Process to account for the black-box nature of $h$;
\item leveraging Expected Improvement to guide Bayesian Optimization;
\item optimizing mixed-type inputs via gradient-based and combinatorial methods.
\end{itemize}





\subsection{Lagrangian Cost Function} \label{sec:31}
As introduced in the previous section, we express the classifier as $h(x) = \mathbb{H}(f(x))$, where $f$ is a real-valued latent function. Since $f$ will later be modeled as a probabilistic surrogate (see Section~\ref{sec:surrogate}), we interpret the decision boundary as the set of points satisfying $f(x) = 0.5$.
Then, the CFE problem~\eqref{eq:optimization} can be written in terms of $f$ as
\begin{equation} \label{eq:penalty}
\underset{x}{\text{minimize}} \ d(x, \tilde{x}) \quad \text{subject to} \ f(x) = 0.5.
\end{equation}
%
In order to solve this problem with Bayesian Optimization, we reformulate it as an unconstrained problem through a Lagrangian formulation.
Let us define the cost function
\begin{align*}
    J(x) = d(x,{\tilde{x}})+\lambda|f(x)-0.5|,
\end{align*}
and let $\{\lambda_k\}_{k=1,2,\dots}$ be a non-negative, increasing sequence tending to infinity. At each iteration $k$, we solve
\begin{equation}\label{eq:EI_penalty}
\underset{x}{\text{minimize}}\quad  d(x,{\tilde{x}})+\lambda_k |f(x)-0.5|,
\end{equation}
by employing an optimization algorithm to identify the minimizer for the current penalty value $\lambda_k$.

As $\lambda_k$ increases, the minimizer of ~\eqref{eq:EI_penalty} will naturally be found in regions where $ |f(x) - 0.5| $ is small. Consequently, as $\lambda_k$ increases, the solutions will progressively move closer to the feasible region $ S_{db} $, and, subject to being close, will minimize $d(x,\tilde{x})$. Ideally, as $\lambda_k \rightarrow \infty$, the solution of problem~\eqref{eq:EI_penalty} is expected to approach the solution of the original constrained problem~\eqref{eq:penalty}. This is the so-called ``penalty function method'' \citep[Sec.~13.1]{luenberger_08}.

The terms $d(x, \tilde{x})$ and $|f(x) - 0.5|$ jointly enforce \emph{proximity}, encouraging counterfactuals that are both close to $\tilde{x}$ and to the decision boundary. Additional constraints are then incorporated to satisfy the key properties for counterfactual explanations discussed in Section~\ref{sec:Intro}.

\vspace{0.5em}
\noindent\textbf{Actionability, Sparsity, and Plausibility}

To enhance interpretability and feasibility, we incorporate additional structural constraints into the optimization problem. Specifically, we enforce the following:

\textit{\textbf{Actionability.}} The counterfactual $x$ must modify only actionable features. We denote by $\mathcal{A} \subseteq \mathcal{X}$ the set of points where immutable features remain unchanged, i.e., $x \in \mathcal{A}$ ensures that $x$ respects domain-specific feasibility constraints such as age or gender~\cite{face_20}.

\textit{\textbf{Sparsity.}} To promote minimal changes, we define a sparsity term $g(x - \tilde{x})$ that penalizes the number or magnitude of feature changes, commonly using norms such as $\ell_0$ or $\ell_1$.

\textit{\textbf{Plausibility.}} To ensure proximity to the data manifold, we use the Local Outlier Factor (LOF)~\cite{breunig_00}. Candidates with LOF scores above a threshold $\tau$ (inliers) receive zero cost, while others (outliers) are discarded via an infinite penalty. We denote this plausibility term as $l(x; X)$, effectively enforcing a hard constraint.

Beyond the structural constraints above, ACE also satisfies key desiderata~\cite{doshi_17} such as validity, diversity, and scalability~\cite{vo_23,Guidotti_22}. Details on how these are addressed by ACE are provided in Appendix~A.
\paragraph{Extended Optimization Problem.} Building upon the three structural constraints introduced above, we define the extended optimization problem as:
\begin{equation}
\label{eq:asce_cost}
\resizebox{0.43\textwidth}{!}{
$
\underbrace{
\underset{x \in \mathcal{A} \subset \mathcal{X}}{\arg\min\;}
}_{\text{actionability}}
\underbrace{d(x, \tilde{x}) + \lambda_k |f(x) - 0.5|}_{\text{proximity}} +
\underbrace{\beta\, g(x - \tilde{x})}_{\text{sparsity}} +
\underbrace{l(x; X)}_{\text{plausibility}}
.
$
}
\end{equation}
As mentioned before, this cost is minimized iteratively by increasing $\lambda_k$ using the penalty method. The hyperparameter $\beta>0$ controls the trade-off with sparsity.

With a slight abuse of notation, we define:
\[
J(x) := d(x, \tilde{x}) + \lambda_k |f(x) - 0.5| + \Theta(x),
\]
where $\Theta(x)$ compactly denotes the additional penalty terms in \eqref{eq:asce_cost} related to sparsity and plausibility.

\subsection{Surrogate Model}\label{sec:surrogate}
To solve optimization problem~\eqref{eq:asce_cost}, we need to approximate $f$ based on samples. However, we may only have access to the classifier output $h(x) \in \{0,1\}$ at a given sample $x \in \mathcal{X}$, rather than the underlying value $f(x)$ required to solve~\eqref{eq:asce_cost}. Therefore, we construct a surrogate function $\hat{f}$—a probabilistic model that estimates the likelihood of Class 1 membership, taking values in $[0,1]$—to approximate $f$.

In this work we employ a Gaussian Process Classifier (GPC)~\citep[Sec.~6.4]{bishop_06} as the surrogate model $\hat{f}(x)$, because it is a non-parametric model, allowing to flexibly capture complex relationships in the data, and it provides measures of uncertainty in its predictions, which is crucial for our Bayesian optimization scheme.

A Gaussian process \citep[Sec.~3.3]{rasmussen_06} $\hat{f}\colon \mathcal{X} \to \mathbb{R}$ is a stochastic process with index set $\mathcal{X}$ such that, for every 
$\boldsymbol{x} = [x_1, \dots, x_n]^T \in \mathcal{X}^n$ (with $n \in \mathbb{N}$ arbitrary), the joint probability density function of $\boldsymbol{\hat{f}} = [\hat{f}(x_1), \dots, \hat{f}(x_n)]^T$ satisfies
\begin{align*}
p(\boldsymbol{\hat{f}} | \boldsymbol{x}) = \mathcal{N}(\boldsymbol{\hat{f}}; \boldsymbol{\mu}, \boldsymbol{K}),
\end{align*}
where $\boldsymbol{\mu} \in \mathbb{R}^n$ and $\boldsymbol{K} \in \mathbb{R}^{n \times n}$ satisfies $K_{ij} = \kappa(x_i, x_j)$ ($i, j \in [n]$), with $\kappa$ being a \emph{kernel} function \citep[Ch.~6]{bishop_06}. In the sequel, we will assume that $\boldsymbol{\mu} = \boldsymbol{0}$. Given observed function values $\boldsymbol{\hat{f}_\ast}$ at points $\boldsymbol{x_\ast} \in \mathcal{X}^n$, suppose we want to predict the value $y = \hat{f}(x)$ at a point $x \in \mathcal{X}$. The joint distribution of the observed values $\boldsymbol{\hat{f}_\ast}$ and the prediction $y$ is also Gaussian:
\begin{align*}
p(y, \boldsymbol{\hat{f}}_\ast | x, \boldsymbol{x}_\ast) = \mathcal{N}\left(\begin{bmatrix}
y \\
\boldsymbol{\hat{f}}_\ast
\end{bmatrix}; \boldsymbol{0}, \begin{bmatrix}
\kappa(x,x) & \boldsymbol{k}_n^T \\
\boldsymbol{k}_n & \boldsymbol{K}
\end{bmatrix} \right),
\end{align*}
where $\boldsymbol{k}_n \in \mathbb{R}^n$ is given by $(k_n)_i = \kappa(x, (x_\ast)_i)$ ($i \in [n]$). From this expression we can derive the conditional density of $y$ given $x$, $\boldsymbol{x}_\ast$ and $\boldsymbol{\hat{f}}_\ast$:
\begin{equation*} 
p(y | x, \boldsymbol{x}_\ast, \boldsymbol{\hat{f}}_\ast) = \mathcal{N}\big( y; \boldsymbol{k}_n^T \boldsymbol{K}^{-1} \boldsymbol{\hat{f}}_\ast, \kappa(x,x)
- \boldsymbol{k}_n^T \boldsymbol{K}^{-1} \boldsymbol{k}_n \big).
\end{equation*}

For classification problems, we aim to estimate the class probability $p(t = 1 | x, \boldsymbol{x}_\ast, \boldsymbol{t}_\ast)$, where $t \in \{0, 1\}$ is the label corresponding to input $x \in \mathcal{X}$, and $\boldsymbol{t}_\ast \in \{0, 1\}^n$ is a vector of labels associated with the training inputs $\boldsymbol{x}_\ast \in \mathcal{X}^n$.

Let $\boldsymbol{a}_\ast = \hat{f}(\boldsymbol{x}_\ast)$ denote the latent function values at the training inputs. The posterior distribution $p(\boldsymbol{a}_\ast | x, \boldsymbol{x}_\ast, \boldsymbol{t}_\ast)$ can be replaced by the Laplace approximation \citep[Sec.~3.4]{rasmussen_06}
\begin{align*}
p(\boldsymbol{a}_\ast | x, \boldsymbol{x}_\ast, \boldsymbol{t}_\ast) \approx \mathcal{N}\left( \boldsymbol{a}_\ast; \hat{\boldsymbol{a}},\; (\boldsymbol{W}(\hat{\boldsymbol{a}})+ \boldsymbol{K}^{-1})^{-1} \right),
\end{align*}
where $\boldsymbol{W}(\boldsymbol{a}) = \text{diag}(\sigma(a_i)[1 - \sigma(a_i)])$ with $\sigma(a) = 1 / (1 + \exp(-a))$ denoting the logistic sigmoid function, and $\hat{\boldsymbol{a}}$ is obtained by iterating until convergence the equation
\begin{equation*}
\hat{\boldsymbol{a}}_{m+1} = \boldsymbol{K} (\boldsymbol{I} + \boldsymbol{W}(\hat{\boldsymbol{a}}_m) \boldsymbol{K})^{-1} (\boldsymbol{t}_\ast - \boldsymbol{\sigma}(\hat{\boldsymbol{a}}_m) + \boldsymbol{W}(\hat{\boldsymbol{a}}_m) \hat{\boldsymbol{a}}_m),
\end{equation*}
with $\boldsymbol{\sigma}(\hat{\boldsymbol{a}}_m)$ denoting the sigmoid applied element-wise.

Given this approximation for the latent values at the training inputs, we can now compute the posterior distribution over the latent function value $a = \hat{f}(x)$ at a test input $x$, which is approximately Gaussian, $\mathcal{N}(\mu_a, \sigma_a^2)$, where $\mu_a$ and $\sigma^2_a$ are the logit mean and the logit variance, respectively, defined as
\begin{align}
\mu_a &= \boldsymbol{k}_n^T (\boldsymbol{t}_\ast - \boldsymbol{\sigma}(\hat{\boldsymbol{a}})), \label{eq:logit_mean} \\
\sigma^2_a &= \kappa(x,x) - \boldsymbol{k}_n^T \left( \boldsymbol{W}(\hat{\boldsymbol{a}})^{-1} + \boldsymbol{K} \right)^{-1} \boldsymbol{k}_n. \label{eq:logit_variance}
\end{align}
The predictive class-1 probability is then approximated using the inverse probit transformation, i.e., $p(t = 1 \mid x, \boldsymbol{x}_\ast, \boldsymbol{t}_\ast) = \sigma\big(\mu_a \big(1+ \frac{\pi\sigma_a^2}{8} \big)^{-1/2} \big).$
See Appendix~C.

%
The posterior mean $\mu_a$ in logit space is transformed into probability via the sigmoid function, yielding $\mu = \sigma(\mu_a)$. The variance in probability space is then $\sigma^2 = \sigma_a^2 \mu^2 (1 - \mu)^2$, obtained via the delta method~\citep[p.~240]{casella_02}, as discussed in Appendix~D.

 \subsection{Expected Improvement}\label{sec:stochastic_EI}
Given the observations and surrogate model, the goal is to decide where to sample next. In Bayesian Optimization, an \emph{acquisition function}---a computationally inexpensive function---estimates the expected gain at each point $x$ and guides the choice of the next query. Ideally, we sample where this value is maximized under the current data.

We use \emph{Expected Improvement} (EI)~\cite{frazier_18} as acquisition function. EI measures the gap between the current optimum and the surrogate function at a point $x \in \mathcal{X}$, i.e.,
\[
\text{EI}_n(x) := \mathbb{E}_n\left\{[J_n^\ast - J(x)]_+\right\} \]
%
where $\mathbb{E}_n$ denotes expectation conditioned on the previously observed data, 
$J(x)$ is defined in Equation~\eqref{eq:asce_cost}, 
and $x^\ast = \arg\min_{x_i \in \boldsymbol{x}_{1:n}} J(x_i)$ is the cost minimizer among the previously observed inputs. 
To estimate $\text{EI}_n(x)$, we use a correlated Monte Carlo sampling method~\citep[Sec.~11.1]{bishop_06}. Accordingly, we define $\hat{g}(x)$ as
\begin{align*}
    \hat{g}(x) &= \text{max}\big( 0, d(x^\ast, \tilde{x}) + \lambda |\hat{f}(x^\ast)-0.5|+\Theta\left(x^*\right) \big.\\
    &\hspace{4.2em} \big. - d(x, \tilde{x}) - \lambda |\hat{f}(x)-0.5|-\Theta(x)\big).
\end{align*}
We then approximate $\text{EI}_n(x)$ as
\vspace{-3pt}
\[
\mathbb{E}_n[ \hat{g}(x) ] \approx \frac{1}{m} \sum_{i=1}^{m} \hat{g}_i(x). \vspace{-3pt}\]
First, let $ \mu(x) $ and $ \sigma^2(x) $ be the posterior mean and variance of the GP model at point $ x $, and similarly for $ x^\ast $. The joint distribution of $ \hat{f}(x) $ and $ \hat{f}(x^\ast) $ is given by:

\begin{center}
\resizebox{0.99\linewidth}{!}{
$\displaystyle
\begin{bmatrix}
\hat{f}(x) \\
\hat{f}(x^\ast)
\end{bmatrix}
\sim \mathcal{N} \left(
\begin{bmatrix}
\mu(x) \\
\mu(x^\ast)
\end{bmatrix},
\ve{\Sigma}_f
\right),
\; \ve{\Sigma}_f =
\begin{bmatrix}
\sigma^2(x) & \text{Cov}(x, x^\ast) \\
\text{Cov}(x, x^\ast) & \sigma^2(x^\ast)
\end{bmatrix}.
$
}
\end{center}
\vspace{2.5pt}
To generate correlated samples, we apply Cholesky decomposition to $ \ve{\Sigma}_f $, i.e., $ \ve{\Sigma}_f = \ve{LL}^\top $, where $\ve{L}$ is a lower triangular matrix. Sampling $ \mathbf{z} \sim \mathcal{N}(\mathbf{0}, \mathbf{I})$, we obtain correlated samples via $ [\hat{f}(x), \hat{f}(x^\ast)]^\top = [\mu(x), \mu(x^\ast)]^\top + \ve{L}\mathbf{z} $. The derivation of the cross-covariance term $\text{Cov}(x, x^\ast)$ is detailed in Appendix~D.

Finally, we approximate the maximizer of \( \text{EI}_n \) using a general-purpose optimization method such as the quasi-Newton L-BFGS-B algorithm~\cite{nocedal_89}.

\subsection{Branch and Bound Method for Mixed Variables}\label{sec:BB}
The ACE algorithm employs a hybrid optimization strategy to handle both continuous and categorical variables. For continuous features, we apply the quasi-Newton L-BFGS-B algorithm to maximize the acquisition function, whereas for categorical variables, which are inherently discrete, ACE employs a Branch and Bound (B\&B) strategy.

Categorical features are encoded using ordinal or label encoding~\cite{sklearn_11}, depending on whether they exhibit a natural order. Starting from the solution obtained via continuous optimization (the \emph{root node}), the B\&B method systematically partitions the categorical space into smaller subproblems by introducing integer constraints. At each branch, the algorithm runs L-BFGS-B over the remaining continuous variables while fixing categorical ones, selecting the candidate that maximizes the acquisition function. For a visual example, see Appendix~E.

\subsection{Overall Algorithm}
The pseudo-code for ACE is outlined in Algorithm \ref{alg:algorithm1}.\\
\textbf{Initialization.} The algorithm starts by generating $n_0$ samples to fit a Gaussian process model. The training data can be obtained either by sampling from a truncated normal distribution — with known mean, variance, and bounds, and uniform sampling for categorical features — or by selecting $n_0$ points from a known training dataset.\\
\textbf{Optimize Acquisition.} After fitting the kernel to the initial dataset $(X, y)$, the algorithm maximizes the acquisition function using the quasi-Newton L-BFGS-B method, starting from a point sampled from a truncated normal centered at $\tilde{x}$. If categorical features are present, the B\&B algorithm is applied, as described in Section~\ref{sec:BB}. For high-dimensional datasets like MNIST, the initial point is drawn from PCA-reduced data~\citep[Sec.~12.1]{bishop_06} and then projected back to the original space before optimization.\\
\noindent\textbf{Filtered Monte-Carlo Expected Improvement.} ACE calculates the EI using Monte Carlo sampling, cf.~Section~\ref{sec:stochastic_EI}, where the number of samples is determined by the parameter \textit{MC}. To evaluate the cost function, both $d(x,\tilde{x})$ and $g(x - \tilde{x})$ from \eqref{eq:asce_cost} are computed using feature-normalized norms—scaling each input dimension by the standard deviation of the corresponding feature in the current dataset $X$—with the $\ell_2$-norm used for proximity and the $\ell_1$-norm for sparsity. Outliers are removed using LOF.\\
\textbf{Best Posterior CFE}. ACE evaluates the posterior mean on a grid of points generated via \textit{Sobol Sampling} \cite{sobol_67}, a low-discrepancy method that ensures a dense and uniform coverage of the search space. This allows identifying points whose posterior probabilities are closest to $0.5$, indicating proximity to the classifier's decision boundary. Among these, the final CFE is selected as the one with the smallest Euclidean distance to $\tilde{x}$. While this step prioritizes proximity, sparsity has already been enforced during optimization via the cost function (cf.~\eqref{eq:asce_cost}), ensuring that the candidate pool reflects both objectives. If the selected candidate ($x_s^n$) has the desired label but is farther from $\tilde{x}$ than the previous CFE found ($x_s^o$), the algorithm terminates; otherwise, the process continues until a closer CFE is identified.
\begin{algorithm}[tb]
   \caption{ACE algorithm}
   \label{alg:algorithm1}

   \textbf{Input}: Initial data $\boldsymbol{n_0}$, instance to explain $\boldsymbol{\tilde{x}}$\\
   \textbf{Parameters}: $\lambda_0$ (init. penalty), $\lambda_{\max}$ (max penalty), $\kappa$ (kernel), $MC$ (MC samples), $SS$ (Sobol samples), $\epsilon$ (tolerance), $p$ (penalty growth) \\
   \textbf{Output}: CFE $\boldsymbol{x_s}$
   \begin{algorithmic}
   \STATE $X,y \gets$ Update Initial Data $(\ve{x}_{1:n_0},h(\ve{x}_{1:n_0}))$
   \WHILE{$h(\tilde{x}) = h(x_s^n)$ \AND $\lVert x_s^n - \tilde{x} \lVert < \lVert x_s^o - \tilde{x} \lVert$}
   \STATE $x_s^o \gets x_s^n$, $\lambda_k \gets \lambda_0$
   \WHILE{$\lVert x_k - x_{k-1} \lVert > \epsilon$ or $\lambda_k < \lambda_{max}$}
    \STATE {$x_k \gets$ maximize EI$_k(X,y,\lambda_k,\kappa,MC)$ over $x$} 
    \STATE Observe $h(x_k)$
    \STATE $X,y \gets$ Update Data $(x_k,h(x_k))$
    \STATE $k \gets k+1$, $\lambda_k \gets (\lambda_{k-1})^p$
   \ENDWHILE
   \STATE $x_s^n \gets$ Sample Decision Boundary $(X,y,\kappa,SS)$
   \ENDWHILE
   \STATE {\bfseries return} $x_s^n$
\end{algorithmic}
\end{algorithm}
\section{Quantitative Evaluation}\label{sec:experiments}
In this section, we define evaluation criteria based on the key properties of CFEs (Section~\ref{sec:31}, Appendix~A) and introduce an aggregated score for comparison. Using these metrics, we benchmark ACE against three state-of-the-art methods across eight binary classification datasets.

\subsection{Experimental Setup}\label{sec:setup}

We evaluate three state-of-the-art methods: BayCon~\cite{baycon_22}, MOC~\cite{moc_20}, and Growing Spheres (GS)~\cite{gs_18}, with the latter implemented using CARLA~\cite{carla_21}. Experiments are conducted on eight real-world classification datasets, detailed in Table~\ref{tab:rw_datasets} and in Appendix~F. Datasets with only numerical features are labeled \textit{continuous}, while those mixing numerical and categorical features are \textit{heterogeneous}.

For each dataset, we perform two evaluations: (i) \emph{fixed-instance}, where a single input instance is selected and each method is executed 100 times using different random seeds to assess robustness and variability; and (ii) \emph{mixed-instance}, which involves generating one counterfactual explanation for each of 100 randomly sampled instances and target labels to evaluate general performance. For BayCon and MOC, which yield multiple candidates, we report the one minimizing the cost in \eqref{eq:asce_cost}, ensuring a fair comparison with ACE and GS, which return a single CFE per run.

Across all experiments, CFEs are generated for a Random Forest black-box model trained via bootstrap aggregation, withholding the instances selected for explanation.

Hyperparameter details are provided in Appendix~G. A key design choice is the kernel used in the Gaussian Process surrogate; we adopt the Matérn 5/2 kernel, as it models $C^2$ functions and better captures complex decision boundaries than RBF—which assumes infinite smoothness—or linear kernels—which impose overly simplistic structure \cite[Sec.~4.2.1]{rasmussen_06}.

All experiments were conducted in Python 3.12.7, using SciPy for optimization\footnote{The GP classifier in ACE was implemented from scratch, as standard libraries (e.g., \texttt{scikit-learn}) do not expose the latent posterior—i.e., the mean and variance in \eqref{eq:logit_mean} and~\eqref{eq:logit_variance}—required to compute covariances between candidates $x$ and the current best $x^\ast$ (see Appendix~D).}.

\begin{table}[tb]
\caption{Summary of Real-world datasets.}
\label{tab:rw_datasets}
\vskip -0.1in
\centering
\scriptsize
\renewcommand{\arraystretch}{0.85}
\setlength{\tabcolsep}{3pt}
\begin{tabular}{lcc}
    \toprule
    Dataset & Features (Numerical/Categorical) & Samples \\
    \midrule
    Diabetes & 8/0 & 768 \\
    KC2 & 21/0 & 522 \\
    Breast Cancer & 9/0 & 683 \\
    Blood & 4/0 & 748 \\
    Tic-Tac-Toe & 0/9 & 958 \\
    Nursery & 0/8 & 12961 \\
    CMC & 2/7 & 1473 \\
    Credit & 4/5 & 1000 \\
    \bottomrule
\end{tabular}
\vskip -0.2in
\end{table}

\subsection{Evaluation Metrics}
The main evaluation metric is the \emph{number of black-box evaluations} $h_\#$, used as a proxy for sample efficiency. We assess \emph{proximity} using the Euclidean distance, defined as $d_2 = \|x - \tilde{x}\|_2$, and \emph{sparsity} using the $\ell_1$ norm, defined as $g_1 = \|x - \tilde{x}\|_1$. In the mixed-instance test setting, both metrics are normalized by the standard deviation of each input feature computed over the entire dataset to account for feature scale differences, and are denoted by $d_{2_N}$ and $g_{1_N}$, respectively.
To measure \emph{plausibility}, we use the affinity score $\alpha(x)$, a smoothed LOF-based metric (cf. Appendix~B) defined as $\alpha(x) := \texttt{clip}\{\exp(1 + \text{LOF}_k(x))\}$, with values close to 1 indicating inliers. We also report \emph{validity} $\mathcal{V}$, the proportion of successful counterfactual generations.

\vspace{1em}
\noindent\textbf{CFE Score}

To compare methods, we define a scalar \emph{CFE Score} $\mathcal{S}$ that aggregates all metrics into a single value using min-max normalization. Since affinity and validity lie in $[0,1]$ and are to be maximized, we use $1 - x_{ij}$ to align interpretation across terms, where lower values are preferred.

Based on the normalized metrics, the CFE Score is defined as:
\[
\mathcal{S} = w_1 \cdot \tilde{h}_\# + w_2 \cdot \tilde{d}_2 + w_3 \cdot \tilde{g}_1 + w_4 \cdot (1 - \alpha) + w_5 \cdot (1 - \mathcal{V}),
\]
where the tilde $\tilde{(\cdot)}$ denotes min-max normalization, and $w_i$ are weights reflecting metric priorities. Since ACE prioritizes sample efficiency, we set $w_1 = 0.35$ for the number of queries $h_\#$. The remaining weights are $0.25$ for $d_2$, $0.15$ for $g_1$, and $0.125$ for both affinity and validity, with $\sum w_i = 1$. A lower $\mathcal{S}$ indicates better performance.


\subsection{Benchmark Results}
\vspace{0.5em} 
\noindent\textbf{Fixed Test Results}

Tables~\ref{tab:fixed_cont} and~\ref{tab:fixed_mix} summarize the evaluation results for continuous and heterogeneous datasets. Each method explains two fixed instances 100 times to assess consistency (mean) and variability (standard deviation in parentheses). Distances for categorical features use consistent encodings (label or ordinal), and identical random seeds ensure reproducibility. Non-actionable features—such as age or gender—are held fixed throughout the optimization to guarantee feasibility and interpretability.

ACE is initialized with $n_0 = 30$ points for all datasets, except for the low-dimensional Blood Test dataset (4 features), where $n_0 = 15$ is used. These initial samples are included in the total evaluation count. Across all eight datasets, ACE consistently outperforms competing methods in terms of the number of black-box evaluations ($h_\#$), successfully generating counterfactuals in every experiment.

\noindent\textbf{\textit{Continuous Datasets}}. On \textit{Diabetes} and \textit{KC2}, ACE achieves the lowest CFE Score, offering a better trade-off across all metrics. It outperforms BayCon and MOC in proximity while maintaining similar sparsity. Although GS obtains slightly better $d_2$, it requires over 28{,}000 queries—compared to around 70 for ACE—making it highly inefficient. On \textit{Breast} and \textit{Blood}, ACE performs best on one instance and is only slightly outperformed by MOC in proximity and sparsity on the other. Notably, ACE consistently achieves $\alpha(x) \approx 1$ and is the only method to produce valid CFEs in all 800 runs.
\begin{table*}[t]
\tiny
\caption{Fixed test for continuous datasets with 100 tested points per experiment. The best Score CFE values are highlighted.}
\label{tab:fixed_cont}
\centering
\setlength{\tabcolsep}{1.5pt}
\renewcommand{\arraystretch}{0.85}
\begin{tabular}{llcccccc ccccccc}
\toprule
\multirow{2}{*}{\textbf{Dataset}} & \multirow{2}{*}{\textbf{Method}} 
& \multicolumn{6}{c}{\textbf{Label 1}} & \multicolumn{6}{c}{\textbf{Label 2}} \\
\cmidrule(lr){3-8} \cmidrule(lr){9-14}
& & \textbf{$h_\#\,$(std)} & \textbf{$d_2\,$(std)} & \textbf{$g_1\,$(std)} & \textbf{$\alpha(x)\,$(std)} & \textbf{$\mathcal{V}$ (std)} & \textbf{$\mathcal{S}$}
  & \textbf{$h_\#\,$(std)} & \textbf{$d_2\,$(std)} & \textbf{$g_1\,$(std)} & \textbf{$\alpha(x)\,$(std)} & \textbf{$\mathcal{V}$ (std)} & \textbf{$\mathcal{S}$} \\
\midrule

\multirow{4}{*}{\textbf{Diabetes}} 
& ACE        & 71 (16) & 34.09 (14.3) & 49.57 (22.34) & 1 (0.06) & 1 (0) & \textbf{0.24} & 70 (15) & 37.10 (6.88) & 70.70 (13.33) & 0.97 (0.04) & 1 (0) & \textbf{0.17} \\
& BayCon      & 1211 (110) & 37.13 (9.23) & 46.87 (12.19) & 0.25 (0.08) & 1 (0) & 0.36 & 1458 (225) & 52.92 (10.5) & 77.56 (14.59) & 0.24 (0.10) & 1 (0) & 0.34 \\
& MOC         & 2038 (1065) & 58.95 (18.46) & 62.91 (16.77) & 0.84 (0.19) & 0.98 (0.14) & 0.45 & 2269 (1247) & 87.00 (70.22) & 98.66 (82.49) & 0.55 (0.16) & 0.98 (0.14) & 0.47 \\
& CARLA$_{GS}$   & 28013 (4455) & 5.51 (0.90) & 12.46 (2.24) & 0.78 (0.03) & 1 (0) & 0.38 & 68937 (7978) & 13.70 (1.60) & 30.20 (4.30) & 0.80 (0) & 1 (0) & 0.38 \\

\midrule

\multirow{4}{*}{\textbf{Breast}} 
& ACE        & 60 (16) & 9.33 (1.05) & 27.89 (2.87) & 1 (0.10) & 1 (0) & \textbf{0.2} & 78 (27) & 15.05 (0.90) & 40.03 (3.24) & 1 (0.01) & 1 (0) & 0.34 \\
& BayCon      & 3567 (1058) & 9.60 (1.25) & 21.12 (1.90) & 0.31 (0.03) & 0.03 (0.20) & 0.35 & 2960 (488) & 13.81 (0.76) & 33.30 (4.61) & 0.29 (0.03) & 0.03 (0.20) & 0.35 \\
& MOC         & 845 (748) & 10.28 (1.19) & 20.81 (3.11) & 0.09 (0.06) & 0.97 (0.17) & 0.37 & 372 (507) & 15.65 (0.78) & 29.97 (3.71) & 0.52 (0.19) & 0.99 (0.10) & \textbf{0.31} \\
& CARLA$_{GS}$   & 40883 (3545) & 9.08 (0.71) & 21.41 (2.25) & 0.77 (0.03) & 1 (0) & 0.39 & 65703 (5583) & 13.03 (1.12) & 32.88 (3.53) & 0.79 (0.02) & 1 (0) & 0.42 \\

\midrule

\multirow{4}{*}{\textbf{KC2}} 
& ACE        & 56 (12) & 626.76 (170.35) & 1436.83 (328.41) & 0.90 (0.22) & 1 (0) & \textbf{0.02} & 57 (13) & 662.05 (170.25) & 1529.80 (322.74) & 0.89 (0.23) & 1 (0) & \textbf{0.12} \\
& BayCon      & 3715 (1498) & 2448.56 (2672.30) & 2843.70 (3083.79) & 0.13 (0.15) & 0.98 (0.14) & 0.14 & 3575 (1177) & 3441.04 (2359.45) & 3988.32 (2674.31) & 0.05 (0.12) & 1 (0) & 0.53 \\
& MOC         & 457 (672) & 62570.15 (165673.31) & 64514.02 (167510.11) & 0.02 (0.08) & 0.19 (0.39) & 0.62 & 1277 (1093) & 1652.42 (1597.79) & 2069.07 (1808.94) & 0.09 (0.19) & 0.13 (0.34) & 0.42 \\
& CARLA$_{GS}$   & 139093 (22089) & 27.73 (4.42) & 101.57 (18.79) & 0.17 (0.02) & 1 (0) & 0.45 & 102423 (42302) & 20.39 (8.45) & 73.65 (30.27) & 0.16 (0.02) & 1 (0) & 0.46 \\

\midrule

\multirow{4}{*}{\textbf{Blood}} 
& ACE        & 54 (15) & 2035.74 (596.52) & 2072.76 (595.77) & 0.99 (0.12) & 1 (0) & \textbf{0.14} & 57 (15) & 78.15 (207.30) & 70.63 (207.07) & 1 (0.02) & 1 (0) & 0.38 \\
& BayCon      & 919 (159) & 2017.03 (104.66) & 2046.13 (107.42) & 0 (0.03) & 1 (0) & 0.31 & 772 (105) & 75.42 (8.98) & 78.77 (10.57) & 0 (0.01) & 1 (0) & 0.6 \\
& MOC         & 1790 (970) & 5999.42 (1407.57) & 6044.28 (1418.61) & 0.05 (0.20) & 0.85 (0.36) & 0.63 & 632 (396) & 8.12 (43.56) & 8.27 (43.94) & 0.99 (0.05) & 0.66 (0.47) & \textbf{0.15} \\
& CARLA$_{GS}$   & 6313 (1239) & 1.15 (0.23) & 1.91 (0.42) & 0.64 (0.16) & 1 (0) & 0.4 & 3083 (271) & 0.55 (0.06) & 0.76 (0.16) & 0.8 (0.03) & 1 (0) & 0.38 \\

\bottomrule
\end{tabular}
\end{table*}
\begin{table*}[t]
\tiny
\caption{Fixed test for heterogeneous datasets with 100 tested points per experiment. The best Score CFE values are highlighted.}
\label{tab:fixed_mix}
\centering
\setlength{\tabcolsep}{1.5pt}
\renewcommand{\arraystretch}{0.85}
\begin{tabular}{llcccccc ccccccc}
\toprule
\multirow{2}{*}{\textbf{Dataset}} & \multirow{2}{*}{\textbf{Method}} 
& \multicolumn{6}{c}{\textbf{Label 1}} & \multicolumn{6}{c}{\textbf{Label 2}} \\
\cmidrule(lr){3-8} \cmidrule(lr){9-14}
& & \textbf{$h_\#\,$(std)} & \textbf{$d_2\,$(std)} & \textbf{$g_1\,$(std)} & \textbf{$\alpha(x)\,$(std)} & \textbf{$\mathcal{V}$ (std)} & \textbf{$\mathcal{S}$}
  & \textbf{$h_\#\,$(std)} & \textbf{$d_2\,$(std)} & \textbf{$g_1\,$(std)} & \textbf{$\alpha(x)\,$(std)} & \textbf{$\mathcal{V}$ (std)} & \textbf{$\mathcal{S}$} \\
\midrule

\multirow{4}{*}{\textbf{CMC}} 
& ACE        & 65 (6) & 1 (0) & 1 (0) & 1 (0.01) & 1 (0) & \textbf{0} & 63 (7) & 1 (0) & 1 (0) & 1 (0.01) & 1 (0) & \textbf{0} \\
& BayCon      & 1021 (102) & 1 (0) & 1 (0) & 0.33 (0.02) & 1 (0) & 0.43 & 925 (46) & 1 (0) & 1 (0) & 0.29 (0.04) & 1 (0) & 0.41 \\
& MOC         & 590 (212) & 1 (0.1) & 1 (0.2) & 0.9 (0.1) & 1 (0) & 0.2 & 528 (184) & 1 (0) & 1 (0) & 0.7 (0) & 1 (0) & 0.21 \\
& CARLA$_{GS}$   & 1005 (0) & 3.87 (0) & 7 (0) & 1 (0) & 1 (0) & 0.74 & 1005 (0) & 4.24 (0) & 8 (0) & 1 (0) & 1 (0) & 0.75 \\

\midrule

\multirow{4}{*}{\textbf{Nursery}} 
& ACE        & 57 (7) & 1 (0.04) & 1.01 (0.1) & 1 (0) & 1 (0) & \textbf{0} & 60 (8) & 1 (0) & 1 (0) & 1 (0.01) & 1 (0) & \textbf{0} \\
& BayCon      & 836 (29) & 1 (0) & 1 (0) & 0.37 (0.01) & 1 (0) & 0.37 & 822 (29) & 1 (0) & 1 (0) & 0.37 (0.01) & 1 (0) & 0.36 \\
& MOC         & 220 (0) & - (-) & - (-) & - (-) & 0 (0) & - & 220 (0) & - (-) & - (-) & - (-) & 0 (0) & - \\
& CARLA$_{GS}$   & 1005 (0) & 3.32 (0) & 5 (0) & 1 (0) & 1 (0) & 0.75 & 1005 (0) & 3.32 (0) & 5 (0) & 1 (0) & 1 (0) & 0.75 \\

\midrule

\multirow{4}{*}{\textbf{German Credit}} 
& ACE        & 68 (26) & 7.48 (9.05) & 12.88 (10.44) & 0.94 (0.08) & 1 (0) & \textbf{0.01} & 54 (10) & 6.11 (2.82) & 11.4 (4.67) & 1 (0.01) & 1 (0) & \textbf{0} \\
& BayCon      & 973 (98) & 1 (0) & 1 (0) & 0.23 (0.05) & 1 (0) & 0.33 & 1013 (69) & 1 (0) & 1 (0) & 0.23 (0.08) & 1 (0) & 0.3 \\
& MOC         & 1436 (753) & 4159.51 (1039.12) & 4200.26 (1047.8) & 0.5 (0.14) & 0.35 (0.48) & 0.89 & 1739 (702) & 3993.84 (473.19) & 4013.8 (472.05) & 0.26 (0.06) & 0.05 (0.22) & 0.96 \\
& CARLA$_{GS}$   & 1005 (0) & 2.13 (0.27) & 3.49 (0.5) & 1 (0.01) & 1 (0) & 0.24 & 1005 (0) & 1.41 (0) & 2.01 (0.01) & 0.99 (0.07) & 1 (0) & 0.2 \\

\midrule

\multirow{4}{*}{\textbf{Tic-Tac-Toe}} 
& ACE        & 67 (11) & 1 (0.01) & 1 (0.03) & 1 (0.01) & 1 (0) & \textbf{0} & 61 (9) & 1 (0.01) & 1 (0.02) & 1 (0.01) & 1 (0) & \textbf{0} \\
& BayCon      & 1526 (274) & 1.41 (0) & 2 (0) & 0.27 (0.02) & 1 (0) & 0.44 & 930 (42) & 1 (0) & 1 (0) & 0.28 (0.02) & 1 (0) & 0.44 \\
& MOC         & 304 (71) & 1.47 (0.17) & 2.19 (0.59) & 0.25 (0.03) & 0.21 (0.41) & 0.59 & 289 (81) & 1.73 (0) & 3 (0) & 0.28 (0.06) & 0.02 (0.14) & 0.7 \\
& CARLA$_{GS}$   & 890111 (312576) & 1.22 (0.25) & 1.55 (0.66) & 1 (0) & 0.11 (0.31) & 0.65 & 1000001 (0) & - (-) & - (-) & - (-) & 0 (0) & - \\

\bottomrule
\end{tabular}
\end{table*}

\noindent\textbf{\textit{Heterogeneous Datasets}}. On \textit{Nursery}, \textit{Tic-Tac-Toe}, and \textit{CMC}, ACE identifies CFEs by changing only one feature by a single unit, while using far fewer evaluations—demonstrating the efficiency of its Branch and Bound strategy. On \textit{German Credit}, ACE achieves the best overall CFE Score despite not being optimal in proximity or sparsity, due to its balance and MOC's poor performance. Once again, ACE maintains high plausibility and validity, confirming its robustness on mixed-variable datasets.

\vspace{0.5em} 
\noindent\textbf{Mixed Test Results}

Tables~\ref{tab:mixed_cont} and~\ref{tab:mixed_mix} present the results of the mixed-instance test, where each method is evaluated on 100 randomly sampled instances and target labels per dataset. A fixed random seed is used throughout to ensure consistency across algorithms, with an initialization setup identical to the fixed-instance test. As in the fixed test, ACE maintains superior sample efficiency, often achieving comparable or better outcomes while requiring significantly fewer evaluations.

\noindent\textbf{\textit{Continuous Datasets}}. ACE achieves the best CFE Score on three of four datasets and ranks second on \textit{Blood}, where BayCon shows marginally better proximity and sparsity. However, BayCon's CFEs are less plausible ($\alpha(x) = 0.32$ vs.\ 0.85 for ACE), indicating they lie farther from the data manifold. ACE also requires only 5.77\% of BayCon's queries on average, highlighting its efficiency.

\noindent\textbf{\textit{Heterogeneous Datasets}}. ACE obtains the best CFE Score on two out of the three datasets, trailing BayCon on \textit{German Credit}. Nevertheless, a similar trade-off is observed: while BayCon achieves slightly better proximity or sparsity, its affinity score is markedly lower ($\alpha(x) = 0.23$ vs.\ 0.94 for ACE). This indicates that ACE generates more feasible and actionable counterfactuals, while operating under far stricter evaluation budgets.
\begin{table}[t]
\tiny
\caption{Continuous Mixed test. Best CFE is highlighted.}
\label{tab:mixed_cont}
\centering
\setlength{\tabcolsep}{0.9pt}
\renewcommand{\arraystretch}{0.85}
\resizebox{\columnwidth}{!}{
\begin{tabular}{llcccccc}
\toprule
\textbf{Dataset} & \textbf{Method} & $h_\#\,$(std) & $d_{2_N}\,$(std) & $g_{1_N}\,$(std) & $\alpha\,$(std) & $\mathcal{V}$ (std) & $\mathcal{S}$ \\
\midrule

\multirow{4}{*}{\textbf{Diabetes}} 
& ACE       & 72 (18) & 2.46 (0.84) & 2.7 (1.87) & 0.98 (0.09) & 1 (0) & \textbf{0.1} \\
& BayCon     & 1239 (238) & 2.04 (0.97) & 2.99 (1.89) & 0.27 (0.09) & 1 (0) & 0.14 \\
& MOC        & 1753 (1011) & 2.09 (0.91) & 2.62 (1.34) & 0.58 (0.36) & 0.91 (0.29) & \textbf{0.1} \\
& CARLA$_{GS}$  & 27463 (23054) & 3.2 (2.66) & 4.26 (3.41) & 0.74 (0.08) & 1 (0) & 0.78 \\

\midrule

\multirow{4}{*}{\textbf{Breast}} 
& ACE       & 67 (22) & 2.45 (1.37) & 7.6 (3.95) & 1 (0.01) & 1 (0) & \textbf{0.16} \\
& BayCon     & 2045 (952) & 2.42 (1.25) & 4.25 (2.63) & 0.30 (0.08) & 0.60 (0.49) & \textbf{0.16} \\
& MOC        & 890 (804) & 3.36 (1.19) & 5.7 (2.51) & 0.47 (0.33) & 1 (0) & 0.39 \\
& CARLA$_{GS}$  & 36803 (20489) & 2.62 (1.46) & 6.56 (3.81) & 0.76 (0.05) & 1 (0) & 0.54 \\

\midrule

\multirow{4}{*}{\textbf{KC2}} 
& ACE       & 57 (13) & 4.05 (1.06) & 14.37 (4.06) & 0.85 (0.23) & 1 (0) & \textbf{0.12} \\
& BayCon     & 3730 (1408) & 2.57 (1.26) & 6.43 (4.01) & 0.14 (0.14) & 0.97 (0.17) & 0.13\\
& MOC        & 1104 (1165) & 11.87 (9.46) & 24.49 (22.74) & 0.12 (0.28) & 0.23 (0.42) & 0.61\\
& CARLA$_{GS}$  & 79763 (39687) & 9.59 (6.57) & 15.67 (8.8) & 0.18 (0.09) & 1 (0) & 0.72 \\

\midrule

\multirow{4}{*}{\textbf{Blood}} 
& ACE       & 63 (23) & 1.83 (0.72) & 2.9 (1.24) & 0.85 (0.32) & 0.99 (0.1) & 0.42 \\
& BayCon     & 1091 (741) & 0.82 (0.65) & 1.02 (0.9) & 0.32 (0.1) & 0.91 (0.29) & \textbf{0.16} \\
& MOC        & 504 (402) & 1.8 (2.75) & 2.48 (4.28) & 0.37 (0.43) & 0.09 (0.29) & 0.56 \\
& CARLA$_{GS}$  & 30633 (28509) & 0.65 (0.7) & 0.85 (0.87) & 0.33 (0.3) & 1 (0) & 0.43 \\

\bottomrule
\end{tabular}}
\vskip -0.15in
\end{table}

\begin{table}[t]
\tiny
\caption{Heterogeneous Mixed test. Best CFE is highlighted.}
\label{tab:mixed_mix}
\centering
\setlength{\tabcolsep}{0.9pt}
\renewcommand{\arraystretch}{0.85}
\resizebox{\columnwidth}{!}{
\begin{tabular}{llcccccc}
\toprule
\textbf{Dataset} & \textbf{Method} & $h_\#\,$(std) & $d_{2_N}\,$(std) & $g_{1_N}\,$(std) & $\alpha\,$(std) & $\mathcal{V}$ (std) & $\mathcal{S}$ \\
\midrule

\multirow{4}{*}{\textbf{CMC}} 
& ACE       & 70 (12) & 2.12 (0.92) & 2.99 (1.8) & 1 (0.01) & 0.97 (0.17) & \textbf{0.07} \\
& BayCon     & 1024 (241) & 1.85 (0.71) & 2.39 (1.24) & 0.32 (0.05) & 0.98 (0.14) & 0.12 \\
& MOC        & 332 (124) & 1.57 (0.68) & 1.8 (1.01) & 0.8 (0.24) & 0.18 (0.38) & 0.13 \\
& CARLA$_{GS}$  & 330674 (469741) & 5.04 (0.9) & 9.19 (1.96) & 1 (0) & 0.67 (0.47) & 0.79 \\

\midrule

\multirow{4}{*}{\textbf{Nursery}} 
& ACE       & 56 (7) & 1.22 (0) & 1.22 (0) & 1 (0.01) & 1 (0) & \textbf{0} \\
& BayCon     & 840 (31) & 1.22 (0) & 1.22 (0) & 0.37 (0) & 1 (0) & 0.37 \\
& MOC        & 226 (10.47) & - (-) & - (-) & - (-) & 0 (0) & - \\
& CARLA$_{GS}$  & 1005 (0) & 2.09 (0.5) & 3.4 (1.27) & 1 (0) & 1 (0) & 0.75 \\

\midrule

\multirow{4}{*}{\makecell[l]{\textbf{German} \\ \textbf{Credit}}}
& ACE       & 73 (24) & 2.61 (1) & 2.47 (2.24) & 0.94 (0.2) & 1 (0) & 0.21 \\
& BayCon     & 1006 (143) & 2.01 (0.57) & 2.11 (0.76) & 0.23 (0.1) & 0.95 (0.22) & \textbf{0.2} \\
& MOC        & 1502 (763) & 2.83 (0.93) & 4.35 (2.02) & 0.27 (0.32) & 0.4 (0.49) & 0.71 \\
& CARLA$_{GS}$  & 3575 (12320) & 2.7 (0.83) & 4.28 (1.6) & 0.95 (0.16) & 1 (0) & 0.71 \\

\bottomrule
\end{tabular}}
\vskip -0.10in
\end{table}

\section{Qualitative Analysis}
While quantitative metrics assess efficiency and quality, visual inspection remains key to understanding counterfactual behavior. We qualitatively analyze ACE’s outputs in both low- and high-dimensional settings, highlighting its coherence and interpretability.
\subsection{Low-Dimensional Visualization}
\begin{figure}[t]
    \centering
    \begin{minipage}{0.475\linewidth}
        \centering
        \includegraphics[width=\linewidth]{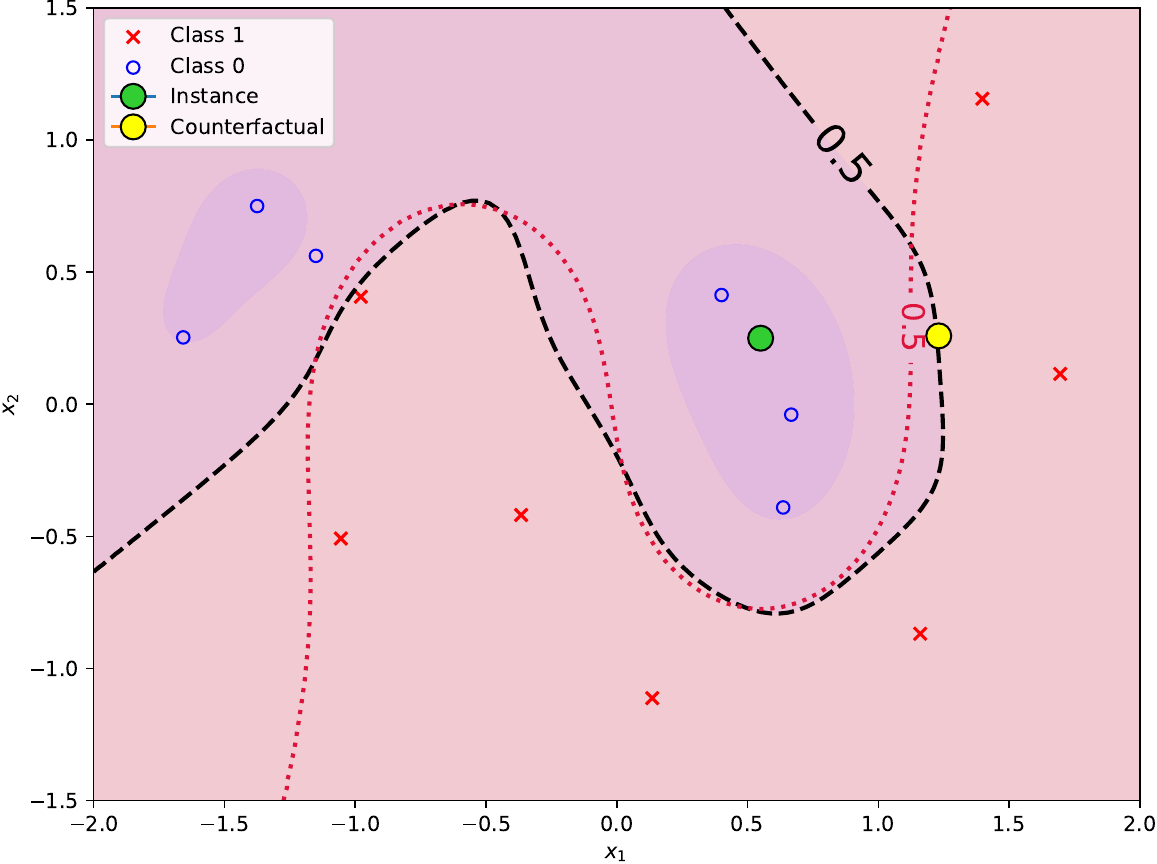}
    \end{minipage}
    \hspace{0.01\linewidth} 
    \begin{minipage}{0.465\linewidth}
        \centering
        \includegraphics[width=\linewidth]{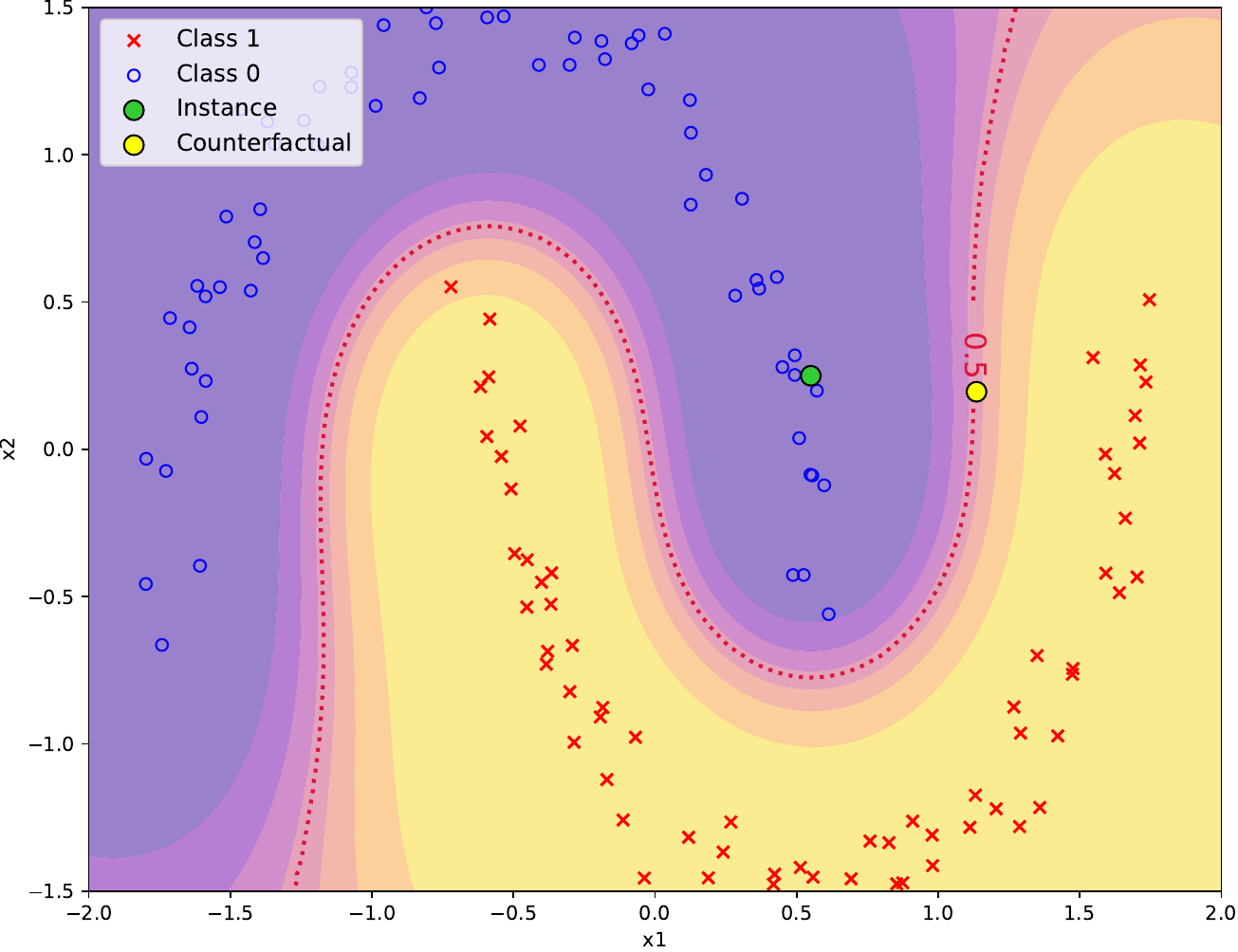}
    \end{minipage}
    \caption{Make Moons Results. From left to right: CFE plot for ACE and CFE plot for GS.}
    \label{fig2}
\vskip -0.08in
\end{figure}
\textit{\textbf{Synthetic Dataset}}. We compare ACE against Growing Spheres (GS)~\cite{laugel_18} on a 2D illustrative example generated via \texttt{make\_moons} from \texttt{scikit-learn}~\cite{sklearn_11}, using a fixed seed for reproducibility. The black-box model is an RBF-kernel SVC with $\gamma = 1$, where $\gamma = 1/(2\sigma^2)$ controls the bias-variance trade-off~\cite[p.~47]{kernels_98}.

Figure~\ref{fig2} compares ACE against GS, which samples within $l_2$-spherical layers until finding an adversarial example, i.e., $h(\text{CFE}_{\text{GS}}) \neq h(\tilde{x})$. The left plot shows the posterior mean $= 0.5$ contour (black dashed), approximating the black-box decision boundary (red dotted). In both plots, the instance $\tilde{x} = [0.55, 0.25]^T$ is marked in green and the resulting CFEs in yellow. With $n_0 = 4$, ACE requires only 14 evaluations to yield $\text{CFE}_{\text{ACE}} = [1.23, 0.25]^T$, whereas GS requires 501 to obtain $\text{CFE}_{\text{GS}} = [1.14, 0.2]^T$. The distances are $d(\tilde{x}, \text{CFE}_{\text{ACE}}) = \mathbf{0.68}$ and $d(\tilde{x}, \text{CFE}_{\text{GS}}) = \mathbf{0.592}$. Notably, ACE moves along a single axis $[+0.68, 0]$, while GS perturbs both axes $[+0.59, -0.05]$, making ACE’s CFE simpler and more efficient despite a slightly larger distance.
\subsection{High-Dimensional Visualization}
\begin{figure}[t]
\begin{center}
\includegraphics[trim=0 4 0 27, clip, width=0.8\columnwidth]{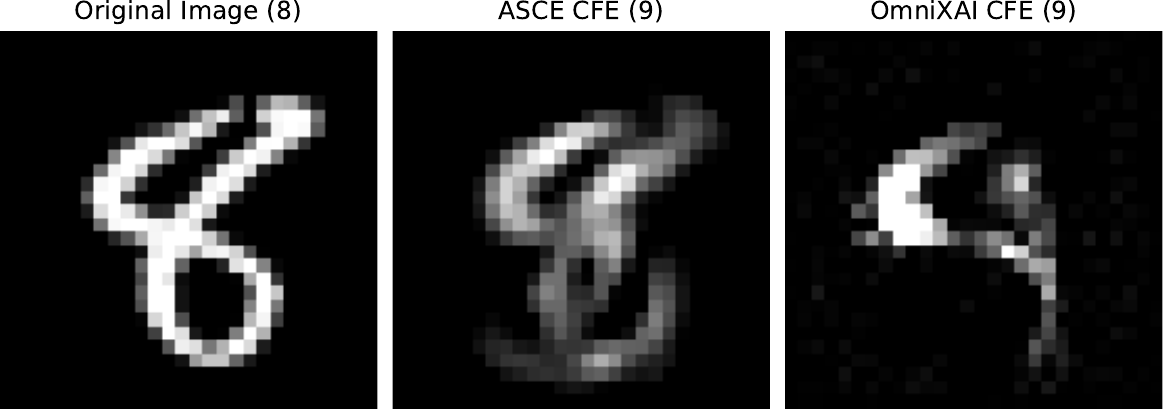} 
\caption{MNIST Results. From left to right: Original instance, ACE counterfactual, and OmniXAI counterfactual.}
\label{fig1}
\end{center}
\vskip -0.25in
\end{figure}
%
\textit{\textbf{MNIST}}. The MNIST dataset~\cite{MNIST_98} consists of grayscale images of handwritten digits (0--9), each represented by a 784-dimensional vector. In our experiment, we focus on digits 8 and 9, and aim to find the minimal Euclidean perturbation that turns a digit 8 into a digit 9. We compare ACE with OmniXAI~\cite{wachter_17}, a library that generates counterfactual examples using a CNN trained specifically for this task. Both methods use the same training data to ensure fairness. The black-box model is a CNN trained for binary classification between 8 and 9, composed of convolutional, pooling, dropout, and dense layers.

In this experiment (Figure~\ref{fig1}), Class~0 corresponds to digit~8 and Class~1 to digit~9. The instance $\tilde{x}$ to be explained (left) belongs to Class~0. ACE, initialized with $n_0 = 50$, finds a valid CFE after only 9 additional queries (59 in total), producing a digit with an open lower loop (center) that visually resembles a 9 and is confidently classified as Class~1 by both the surrogate and black-box models. OmniXAI fails with 50 queries but succeeds after retraining on the same 59 points (right). ACE achieves an $\ell_2$ distance of 5.47 and a posterior probability of 55.19\%, indicating a counterfactual closer to the decision boundary (50\%) compared to OmniXAI’s 7.44 and 64.93\%. Overall, ACE demonstrates superior sample efficiency and plausibility, yielding concise and actionable explanations.
\section{Conclusion}
We propose the Adaptive sampling for Counterfactual Explanations (ACE) algorithm, designed to generate reliable and precise CFEs in a sample-efficient manner. Across real-world and synthetic datasets, ACE outperforms state-of-the-art methods by requiring fewer black-box evaluations while producing meaningful explanations. We envision ACE as a step toward deployable XAI systems in domains where queries are costly, aligning with emerging AI regulations and practical needs. Future work includes multi-objective diversity schemes and theoretical study of ACE’s properties.

\bibliography{aaai2026}
\pagebreak
\appendix
\section*{Appendices}
\section{Desiderata for Counterfactual Explanations}\label{sec:appA}
As identified in prior literature~\cite{vo_23,Guidotti_22}, high-quality counterfactual explanations (CFEs) should satisfy the following key properties, which are effectively addressed by our proposed method:\\
\textbf{Validity.} \textit{The counterfactual must change the predicted label.} ACE guarantees validity by explicitly searching for counterfactuals $x$ such that the predicted outcome differs from the original instance $\tilde{x}$, i.e., $h(x) \neq h(\tilde{x})$.\\
\textbf{Sparsity.} \textit{Minimal number of features should be altered.} To encourage sparse solutions, ACE incorporates an $\ell_1$-norm penalty in its cost function, promoting counterfactuals that modify as few features as possible.\\
\textbf{Proximity.} \textit{Counterfactuals should be close to the original input.} ACE measures proximity via a distance metric $d(x, \tilde{x})$, typically the Euclidean norm, though other metrics can be used depending on the data characteristics. This ensures that explanations remain within a small, interpretable neighborhood $\varepsilon$ around $\tilde{x}$, i.e., $d(x, \tilde{x}) < \varepsilon$.\\
\textbf{Actionability.} \textit{Only mutable features should be changed.} ACE supports user-defined feature constraints and restricts modifications to actionable features, leaving immutable attributes (e.g., age or gender) unchanged throughout the optimization process to ensure the generation of feasible CFEs.\\
\textbf{Diversity.} \textit{Providing multiple distinct counterfactuals improves user choice.} ACE combines Monte Carlo sampling with GP uncertainty to explore the decision boundary globally, while prioritizing diverse candidates near $\tilde{x}$. This dual mechanism enables the generation of multiple, semantically distinct yet plausible CFEs.\\
\textbf{Plausibility.} \textit{Counterfactuals should respect feature constraints and data distribution.} ACE enforces domain constraints and avoids unrealistic combinations by sampling within the input domain and promoting CFEs in high-density regions with respect to the training data.\\
\textbf{Scalability.} \textit{Efficient generation across multiple instances.} Thanks to its Bayesian surrogate model, ACE reuses learned structures across similar queries and scales to both low- and high-dimensional datasets, supporting simultaneous generation of multiple CFEs.
\section{Extended Cost Function}
\textit{\textbf{Plausibility Term.}} To ensure closeness to the data manifold, we penalize counterfactuals that lie in low-density regions using the \emph{Local Outlier Factor (LOF)}~\cite{breunig_00}. LOF quantifies the degree to which a point is isolated from its neighbors, with scores typically interpreted as
\begin{equation}\label{LOF}
\text{LOF}_k(x) := \frac{1}{|N_k(x)|} \sum_{z \in N_k(x)} \frac{\text{lrd}(z)}{\text{lrd}(x)},
\end{equation}
where $N_k(x)$ is the $k$-nearest neighborhood of $x$, and $\text{lrd}(z)$ denotes the local reachability density. In our implementation, we adopt \texttt{scikit-learn}'s \texttt{LocalOutlierFactor} with \texttt{novelty=True}, which outputs negative LOF scores. A point $x$ is considered an \emph{inlier} if its LOF score exceeds a threshold $\tau$ (typically $\tau = -1.5$); otherwise, it is deemed implausible. Thus, we define the penalty
\begin{equation*}
l(x; X) := 
\begin{cases}
0, & \text{if } \text{LOF}_k(x) > \tau \ (\text{inlier}), \\
\infty, & \text{otherwise (outlier)},
\end{cases}
\end{equation*}
which effectively acts as a hard constraint that discards implausible candidates.
\section{Class-1 Posterior Calculations}

For classification problems, we aim to estimate the probability $p(t = 1 | x, \boldsymbol{x}_\ast, \boldsymbol{t}_\ast)$, where $l \in \{0, 1\}$ is the label corresponding to input $x \in \mathcal{X}$, and $\boldsymbol{t}_\ast \in \{0, 1\}^n$ is a vector of labels at the inputs in $\boldsymbol{x}_\ast \in \mathcal{X}^n$. This probability will be approximated as
\begin{align*}
p(t = 1 | x, \boldsymbol{x}_\ast, \boldsymbol{t}_\ast) = \int p(t = 1 | a) p(a | x, \boldsymbol{x}_\ast, \boldsymbol{t}_\ast) da,
\end{align*}
where $a = \hat{f}(x)$, $p(t = 1 | a) = \sigma(a) := 1/(1 + \exp(-a))$ is the logistic sigmoid function, and $\hat{f}$ is a Gaussian process. We are actually interested only on $p(a | x, \boldsymbol{x}_\ast, \boldsymbol{t}_\ast)$, given by 
\begin{align*}
p(a | x, \boldsymbol{x}_\ast, \boldsymbol{t}_\ast) = \int p(a | x, \boldsymbol{x}_\ast, \boldsymbol{a}_\ast) p(\boldsymbol{a}_\ast | x, \boldsymbol{x}_\ast, \boldsymbol{t}_\ast) d \boldsymbol{a}_\ast,
\end{align*}
where $p(a | x, \boldsymbol{x}_\ast, \boldsymbol{a}_\ast)$ is a Gaussian distribution given by
\begin{equation*} 
p(a | x, \boldsymbol{x}_\ast,  \boldsymbol{a}_\ast) = \mathcal{N}\big( a; \boldsymbol{k}_n^T \boldsymbol{K}^{-1}  \boldsymbol{a}_\ast, \kappa(x,x)
- \boldsymbol{k}_n^T \boldsymbol{K}^{-1} \boldsymbol{k}_n \big).
\end{equation*} The density $p(\boldsymbol{a}_\ast | x, \boldsymbol{x}_\ast, \boldsymbol{t}_\ast)$, on the other hand, can be replaced by the Laplace approximation \citep[Sec.~3.4]{rasmussen_06}
\begin{align*}
p(\boldsymbol{a}_\ast | x, \boldsymbol{x}_\ast, \boldsymbol{t}_\ast) \approx \mathcal{N}\left( \boldsymbol{a}_\ast; \hat{\boldsymbol{a}},\; (\boldsymbol{W}(\hat{\boldsymbol{a}})+ \boldsymbol{K}^{-1})^{-1} \right),
\end{align*}
where $\boldsymbol{W}(\boldsymbol{a}) = \text{diag}(\sigma(a_i)[1 - \sigma(a_i)])$, and $\hat{\boldsymbol{a}}$ is obtained by iterating until convergence the equation
\begin{equation*}
\hat{\boldsymbol{a}}_{m+1} = \boldsymbol{K} (\boldsymbol{I} + \boldsymbol{W}(\hat{\boldsymbol{a}}_m) \boldsymbol{K})^{-1} (\boldsymbol{t}_\ast - \boldsymbol{\sigma}(\hat{\boldsymbol{a}}_m) + \boldsymbol{W}(\hat{\boldsymbol{a}}_m) \hat{\boldsymbol{a}}_m),
\end{equation*}
where $\boldsymbol{\sigma}$ consists in the entry-wise application of $\sigma$. Finally,
\begin{align*}
p(a | x, \boldsymbol{x}_\ast, \boldsymbol{t}_\ast) \approx \mathcal{N}\left( a;\; \mu_a,\; \sigma^2_a \right),
\end{align*}
where $\mu_a$ and $\sigma^2_a$ are the logit mean and the logit variance, respectively, defined as
\begin{align}
\mu_a &= \boldsymbol{k}_n^T (\boldsymbol{t}_\ast - \boldsymbol{\sigma}(\hat{\boldsymbol{a}})), \label{eq:logit_mean} \\
\sigma^2_a &= \kappa(x,x) - \boldsymbol{k}_n^T (\boldsymbol{W}(\hat{\boldsymbol{a}})^{-1} + \boldsymbol{K})^{-1} \boldsymbol{k}_n. \label{eq:logit_variance}
\end{align}
The Class 1 probability at input $x$ given $\boldsymbol{x_\ast}$ and $\boldsymbol{t_\ast}$ is calculated using the inverse probit function $\sigma(a) \simeq \Phi(\lambda a)$, obtaining the approximate predictive distribution of the form
\begin{align*}
p(t = 1 | x, \boldsymbol{x}_\ast, \boldsymbol{t}_\ast) = \sigma\big(\mu_a \big(1+ \frac{\pi\sigma_a^2}{8} \big)^{-1/2} \big).
\end{align*}
\section{Delta Method for $\text{Cov}(x, x^\ast)$}
To calculate the covariance between $x$ and $x^\ast$, we apply the delta method which uses the first-order Taylor expansion to approximate the expectation of a function of random variables, in particular, of $g(f(x))$ and $g(f(x^\ast))$, where $g$ is a nonlinear function. Using the first-order Taylor expansion around the logit means $\mu_a(x)$ and $\mu_a({x}^\ast)$ (cf.~\eqref{eq:logit_mean}) yields
\begin{align*}
   &g(f(x)) \approx g(\mu_a(x)) + g'(\mu_a(x))(\hat{f}(x) - \mu_a(x)), \\
   &g(f(x^\ast)) \approx g(\mu_a({x}^\ast)) + g'(\mu_a({x}^\ast))(\hat{f}(x^\ast) - \mu_a({x}^\ast)).
\end{align*}
The covariance is then approximated by
\begin{align*}
\text{Cov}&(g(\hat{f}(x)), g(\hat{f}(x^\ast))) \\
&\approx \text{Cov}\big(g(\mu_a(x)) + g'(\mu_a(x))(\hat{f}(x) - \mu_a(x)), \\
&\hspace{3.4em} g(\mu_a(x^\ast)) + g'(\mu_a(x^\ast))(\hat{f}(x^\ast) - \mu_a(x^\ast))\big) \\
&= g'(\mu_a(x)) g'(\mu_a(x^\ast)) \, \text{Cov}(\hat{f}(x), \hat{f}(x^\ast)),
\end{align*}
where $\text{Cov}((\hat{f}(x), \hat{f}(x^\ast))$ is calculated using  (\ref{eq:logit_variance}) as
\begin{align*}
&\text{Cov}((\hat{f}(x), \hat{f}(x^\ast)) \\&= \kappa(x,x^\ast) -  \kappa(\boldsymbol{x}_{1:n},x)^T (\boldsymbol{W}(\hat{\boldsymbol{a}})^{-1} + \boldsymbol{K})^{-1} \kappa(\boldsymbol{x}_{1:n},x).
\end{align*}
For GPC, $g(z)$ is represented by the sigmoid function $\sigma$, converting logits into probabilities. Consequently, the derivatives are given by $g'(z) = \sigma(z)(1 - \sigma(z))$. Finally,
\begin{align*}
&\text{Cov}(\sigma(\hat{f}(x)), \sigma(\hat{f}(x^\ast)))\\
&\approx \mu_x(1- \mu_x)\mu_{x^\ast}(1-\mu_{x^\ast}) \, \text{Cov}(\hat{f}(x), \hat{f}(x^\ast)) \\
&\approx \text{Cov}(x,x^\ast).
\end{align*}
\section{Branch and Bound Example}
\begin{figure}
    \begin{center}
    \includegraphics[width=0.8\linewidth]{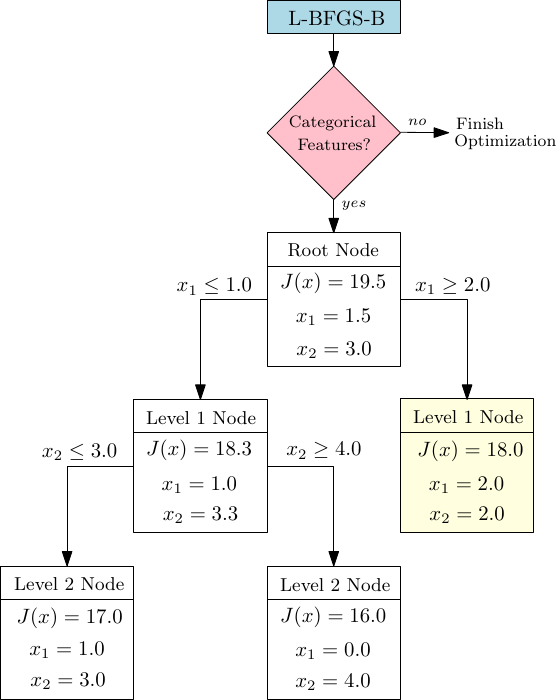}
    \caption{Branch and Bound method with L-BFGS-B Initialization.}
    \label{fig:BB}
    \end{center}
    \vskip -0.25in
\end{figure}

Figure~\ref{fig:BB} illustrates the Branch and Bound algorithm for two categorical variables. Each node represents a constrained maximization, and the yellow-highlighted box corresponds to the best candidate found ($J(x)=18.0$, $x_1=2.0$, $x_2=2.0$). This structured refinement ensures a principled and efficient exploration of the solution space, enabling the identification of near-global optima in problems involving discrete variables and non-convex structures, while effectively pruning suboptimal branches.

\section{Datasets}\label{sec:appA}

The quantitative evaluation is conducted on eight real-world classification datasets sourced from OpenML~\cite{openml}, Kaggle~\cite{kaggle}, and the UCI ML Repository~\cite{uci_ml_repository}. The selected datasets largely align with those used in the original evaluations of the compared methods. Growing Spheres was evaluated on MNIST (in the original paper) and on Make Moons (illustrative examples provided in its public repository). For the eight benchmark datasets, we included those used in the MOC and BayCon studies, which also compared against each other, and added Growing Spheres to enable both qualitative and quantitative comparisons. For OmniXAI, the MNIST dataset was included to complement the visual comparison with Growing Spheres.

The websites corresponding to the datasets used in Section~4 are listed below:

\noindent \begin{links} 
\link{Diabetes}{https://www.openml.org/d/37}
\link{Breast}{https://www.openml.org/d/15}
\link{Blood}{https://www.openml.org/d/1464}
\link{KC2}{https://www.kaggle.com/datasets/chaitanyasirivuri/kc2-software-fault-prediction-dataset}
\link{Tic-Tac-Toe}{https://www.kaggle.com/datasets/rsrishav/tictactoe-endgame-data-set}
\link{Nursery}{https://archive.ics.uci.edu/dataset/76/nursery}
\link{CMC}{https://archive.ics.uci.edu/dataset/30/contraceptive+method+choice}
\link{German Credit}{ttps://www.kaggle.com/datasets/uciml/german-credit}
\end{links} 
\section{Implementation Details}\label{sec:appB}

\subsection{Hyperparameter Search}
An exhaustive grid search is performed over the ranges in Table~\ref{tab:hyperparameters_search}. Continuous ranges (``to'') are sampled at regular intervals, and bracketed values denote discrete candidate sets. Final values are selected from the subrange where further changes have negligible impact on the generated CFE, ensuring robustness and efficiency.

\begingroup
\renewcommand{\arraystretch}{1.1} 
\begin{table}[h]
\caption{Hyperparameter ranges explored.}
\label{tab:hyperparameters_search}
\vskip 0.1in
\begin{center}
\begin{footnotesize}
\begin{tabular}{@{}lcc@{}}
\toprule
\textbf{Name} & \textbf{Symbol} & \textbf{Range Tried} \\
\midrule
Initial Penalty & $\lambda_0$ & 2 to 12 \\
Kernel ($\nu$) & $\kappa$ & $[\tfrac{1}{2}, \tfrac{3}{2}, \tfrac{5}{2}]$ \\
Monte-Carlo Samples & $MC$ & 800 to 2000 \\
Penalty Growth & $p$ & 1.1 to 1.9 \\
Sobol Samples & $SS$ & 1000 to 10000 \\
Sparsity Trade-off & $\beta$ & $[0.1, 1, 3, 5, 7, 10, 100]$ \\
\bottomrule
\end{tabular}
\end{footnotesize}
\end{center}
\vskip -0.2in
\end{table}
\endgroup

As noted above, we chose hyperparameters so that small perturbations would not materially change the resulting CFE, ensuring stable convergence within a well-defined region; the final values used in our experiments are reported in Table~\ref{tab:hyperparameters}. The algorithm starts with an initial penalty value, $\lambda_0$, in the vicinity of 1---set to $10$ in our runs---ensuring that the search begins near the instance to be explained and helps the algorithm efficiently find a nearby minimizer. After each acquisition function maximization, $\lambda$ is scaled by $p = 1.5$, gradually increasing to guide the search toward the decision boundary. Sobol sampling ($SS = 8000$) is utilized to densely cover high-dimensional spaces, ensuring thorough exploration across all features in the dataset, while Monte Carlo sampling ($MC = 1000$) provides an efficient approximation of the Expected Improvement (EI) integral, balancing computational cost and precision.
\begin{table}[h]
\caption{Summary of Hyperparameters.}
\label{tab:hyperparameters}
\vskip 0.1in 
\begin{center}
\begin{footnotesize} 
\begin{tabular}{@{}lcc@{}}
    \toprule
    \textbf{Name} & \textbf{Symbol} & \textbf{Value} \\
    \midrule
    Initial Penalty & $\lambda_0$ & $10$ \\
    Maximum Penalty & $\lambda_{max}$ & $1\mathrm{e}15$ \\
    Kernel & $\kappa$ & Matern $\frac{5}{2}$ \\
    Monte-Carlo Samples & $MC$ & $1000$ \\
    Convergence Tolerance & $\epsilon$ & $0.001$ \\
    Penalty Growth & $p$ & $1.5$ \\
    Sobol Samples & $SS$ & $8000$ \\
    Sparsity trade-off parameter & $\beta$ & $5$ \\
    \bottomrule
\end{tabular}
\end{footnotesize}
\end{center}
\vskip -0.2in 
\end{table}
\subsection{Computing Infrastructure}
All experiments are conducted on a system with an \textbf{AMD Ryzen 7 4800HS CPU @ 2.90\,GHz}, \textbf{16\,GB RAM}, running \textbf{Windows 11 Home Single Language}, with computation performed on CPU only and no GPU acceleration (integrated or discrete) used. The implementation is in \textbf{Python 3.12.7} with the following key libraries: NumPy~(v1.26.4), SciPy~(v1.13.1), scikit-learn~(v1.5.1), pandas~(v2.2.2), matplotlib~(v3.9.2), scikit-image~(v0.24.0), and threadpoolctl~(v3.5.0).

\section{ACE Code}
\subsection{Gaussian Process Surrogate}

The following code implements the Gaussian Process surrogate model used within ACE. 
It includes functions for training a Gaussian Process Classifier with a Matérn kernel, 
computing the weight matrix for the Laplace approximation, iteratively estimating the 
latent vector via Newton–Raphson updates, and computing posterior predictions (mean and 
variance) for new samples.\\

\lstset{
  basicstyle=\ttfamily\scriptsize,
  breaklines=true,
  columns=fullflexible,
  keepspaces=true,
  showstringspaces=false,
  upquote=true
}
\begin{lstlisting}[language=Python]
def train_kernel(X, t, opt, length_scale=1, tol=1e-15):
    """
    Train a GP classifier with Matern 5/2 kernel. Returns (model, fitted_kernel).
    """
    if len(np.unique(t)) < 2:
        print("Only one class present. Skipping.")
        return None, None
    kernel = Matern(length_scale=length_scale, nu=2.5)
    if opt:
        model = GaussianProcessClassifier(kernel=kernel, optimizer='fmin_l_bfgs_b')
    else:
        model = GaussianProcessClassifier(kernel=kernel, optimizer=None)
    model.fit(X, t)
    return model, model.kernel_

def W(a):
    """
    Diagonal weight matrix for Laplace: sigma(a)*(1-sigma(a)).
    """
    sig = sigmoid(a) * (1 - sigmoid(a))
    return np.diag(sig.ravel())

def a_t(X, t, K_a, max_iter=10, tol=1e-6):
    """
    Newton--Raphson refinement of latent vector a.
    """
    a = np.zeros_like(t)
    I = np.eye(X.shape[0])
    for _ in range(max_iter):
        W_a = W(a)
        F1 = np.linalg.inv(I + W_a @ K_a)
        a_new = (K_a @ F1) @ (t - sigmoid(a) + W_a @ a)
        if np.linalg.norm(a_new - a) < tol:
            a = a_new
            break
        a = a_new
    return a

def posterior(X, t, X2, kernel):
    """
    Laplace GP classification posterior: mean/var at X2 (probit approx).
    """
    K   = kernel(X,  X)
    a   = a_t(X, t, K)
    Ks  = kernel(X,  X2)
    Kss = kernel(X2, X2)

    W_inv = np.linalg.inv(W(a))
    F1    = np.linalg.inv(W_inv + K)

    mu  = Ks.T @ (t - sigmoid(a))
    var = np.diag(Kss).reshape(-1,1) - np.sum((F1 @ Ks) * Ks, axis=0).reshape(-1,1)

    # Probit approximation and variance propagation
    kappa   = 1.0 / np.sqrt(1.0 + np.pi * var / 8)
    mu_real = sigmoid(kappa * mu)
    var_real = var * (mu_real * (1 - mu_real))**2
    return mu_real, var_real, F1
\end{lstlisting}

\subsection{Optimization and Expected Improvement}

This component maximizes the acquisition function using multi-start L-BFGS-B combined with a greedy branch-and-bound refinement for categorical features. Expected improvement is computed via Monte Carlo with a cost that balances proximity, sparsity, and 
boundary penalties.\\
\lstset{
  basicstyle=\ttfamily\scriptsize,
  breaklines=true,
  columns=fullflexible,
  keepspaces=true,
  showstringspaces=false,
  upquote=true
}

\begin{lstlisting}[language=Python]
def optimize_acquisition_bb2(
 X, t, categorical_columns, X_test, kernel, bound_vals,
 x_s, MC, factor, lambd=10, n_neighbors=20, action=None,
 sampling_method='lhs', gtol=1e-20):
    """
    Maximize EI under mixed inputs using L-BFGS-B root + greedy BnB.
    Returns best_x, fx, best_ei, x_min.
    """
    def objective(x):
        x = x.reshape(1, -1)
        return -expected_improvement_mc(x, X, t, kernel, x_s, lambd, MC)[0]
        # or: return -expected_improvement_mc_l1(...)[0]

    effective_cats = [c for c in categorical_columns if action is None or c not in action]
    best_result, best_x = None, None
    unique_points = {tuple(y) for y in X}
    unique_points.add(tuple(x_s[0]))

    # Multi-start L-BFGS-B to get a strong root
    for _ in range(10):
        if sampling_method == 'lhs':
            init = latin_hypercube_sample(bound_vals, 1)[0]
        elif sampling_method == 'normal':
            std = np.sqrt(np.abs(x_s.ravel()))
            init = truncated_normal(x_s.ravel(), std, bound_vals[:,0], bound_vals[:,1], 1, factor).ravel()
        elif sampling_method == 'random':
            init = np.random.uniform(bound_vals[:,0], bound_vals[:,1])
        elif sampling_method == 'test':
            init = X_test[np.random.choice(X_test.shape[0])]
            while tuple(init) in unique_points:
                init = X_test[np.random.choice(X_test.shape[0])]
        else:
            init = np.random.uniform(bound_vals[:,0], bound_vals[:,1])

        if tuple(init) in unique_points:
            continue

        res = minimize(objective, init, method='L-BFGS-B',
                       bounds=bound_vals, options={'gtol': gtol})
        if -res.fun <= 0:
            continue

        if best_result is None or res.fun < best_result:
            if filter_outliers(res.x, X, n_neighbors):
                root = res.x.copy()
                # reset so BnB compares only feasible integral solutions
                best_result, best_x = None, None

                def branch_and_bound(curr_point, curr_bounds, level):
                    nonlocal best_result, best_x
                    if level == len(effective_cats):
                        rr = minimize(objective, curr_point, method='L-BFGS-B',
                                      bounds=curr_bounds, options={'gtol': gtol})
                        imp = -rr.fun
                        if best_result is None or imp > -best_result:
                            best_result, best_x = -imp, rr.x
                        return
                    col = effective_cats[level]
                    lo = int(np.floor(curr_point[col]))
                    hi = int(np.ceil(curr_point[col]))
                    for val in range(lo, hi + 1):
                        mp = curr_point.copy()
                        mb = curr_bounds.copy()
                        mp[col] = val
                        mb[col,:] = [val, val]
                        rr = minimize(objective, mp, method='L-BFGS-B',
                                      bounds=mb, options={'gtol': gtol})
                        branch_and_bound(rr.x, mb, level + 1)

                branch_and_bound(root, bound_vals.copy(), 0)
                break

    if best_x is None:
        return None, 0.0, None, 0
    _, fx, x_min = expected_improvement_mc(best_x.reshape(1,-1), X, t, kernel, x_s, lambd, MC)
    return best_x, fx, -best_result, x_min

def expected_improvement_mc_l1(X2, X, t, kernel, x_s, lambda_, n_samples, alpha=5):
    """
    Monte Carlo EI with correlated coupling; cost = d2 + alpha*l1 + lambda*|0.5 - f|.
    Returns mean_improvement, fx_at_argmax, x_min.
    """
    mu_tr, sig_tr, F1 = posterior(X, t, X, kernel)
    mu_st, sig_st, _  = posterior(X, t, X2, kernel)
    sdev = np.std(X, axis=0)

    d  = feature_normalized_distance(X,  x_s, sdev)
    g  = feature_normalized_l1_distance(X, x_s, sdev)
    min_idx = np.argmin(d + alpha*g + lambda_ * np.abs(0.5 - mu_tr))
    x_min   = X[min_idx].reshape(1, -1)
    mu_min  = mu_tr[min_idx]
    var_min = sig_tr[min_idx]

    Ksm = kernel(X2, x_min); KsX = kernel(X2, X); KmX = kernel(X, x_min)
    cov_star_min = Ksm - KsX @ F1 @ KmX
    covar_real   = cov_star_min * (mu_st*(1 - mu_st)) * (mu_min*(1 - mu_min))

    mu_c = np.hstack([mu_st.ravel(), mu_min.ravel()])
    cov_c = np.block([[sig_st,      covar_real     ],
                      [covar_real.T, var_min.reshape(1,1)]])

    samples = sample_gp_correlated(mu_c, cov_c, n_samples)
    f_star = samples[:, 0]; f_min = samples[:, 1]

    d_s   = feature_normalized_distance(X2,   x_s, sdev)
    d_min = feature_normalized_distance(x_min, x_s, sdev)
    g_s   = feature_normalized_l1_distance(X2,   x_s, sdev)
    g_min = feature_normalized_l1_distance(x_min, x_s, sdev)

    c_s   = d_s.reshape(-1,1) + alpha*g_s.reshape(-1,1) + lambda_ * np.abs(0.5 - f_star.reshape(-1,1))
    c_min = d_min.reshape(-1,1) + alpha*g_min.reshape(-1,1) + lambda_ * np.abs(0.5 - f_min.reshape(-1,1))

    improv = np.maximum(0, c_min - c_s)
    mean_improv = np.mean(improv)
    fx = f_star[np.argmax(improv)]
    return mean_improv, fx, x_min

def sample_gp_correlated(mu, cov, n_samples, jitter=1e-6):
    """
    Draw samples from N(mu, cov); add jitter to diag if needed for stability.
    Returns (n_samples, len(mu)).
    """
    try:
        L = np.linalg.cholesky(cov)
    except np.linalg.LinAlgError:
        cov = cov + np.eye(cov.shape[0]) * jitter
        L = np.linalg.cholesky(cov)
    z = np.random.normal(size=(n_samples, len(mu)))
    return mu + z @ L.T

def truncated_normal(mean, std_dev, lower_bound, upper_bound, size,
                     sampling_factor=1.0, min_std=1e-3):
    """
    Draw from truncated Normal with per-dimension bounds; respects fixed dims where lower==upper.
    """
    adjusted_std = np.maximum(std_dev * sampling_factor, min_std)
    samples = np.zeros((size, len(mean)))
    for i in range(size):
        sample = np.zeros(len(mean))
        for j in range(len(mean)):
            if lower_bound[j] == upper_bound[j] or adjusted_std[j] == 0:
                sample[j] = mean[j]
            else:
                while True:
                    a = (lower_bound[j] - mean[j]) / adjusted_std[j]
                    b = (upper_bound[j] - mean[j]) / adjusted_std[j]
                    sample[j] = truncnorm(a, b, loc=mean[j], scale=adjusted_std[j]).rvs(1)[0]
                    if lower_bound[j] <= sample[j] <= upper_bound[j]:
                        break
        samples[i] = sample
    return samples

def sobol_sample(bounds, n_points=100, categorical_columns=None, action=None, seed=None):
    """
    Sobol' sampling with mixed variables:
    - continuous dims: scale to [lower, upper]
    - categorical dims: snap to integer grid
    - action: indices to freeze (lower==upper)
    Returns (n_points, n_dims).
    """
    n_dim = len(bounds)
    sampler = qmc.Sobol(d=n_dim, scramble=True, seed=seed)
    m = int(np.ceil(np.log2(n_points)))  # use 2^m points, then trim
    base = sampler.random_base2(m=m)[:n_points]
    scaled = np.zeros_like(base)
    for i, (lower, upper) in enumerate(bounds):
        if action and i in action:
            scaled[:, i] = lower
        elif categorical_columns and i in categorical_columns:
            values = np.arange(lower, upper + 1)
            idx = np.round(base[:, i] * (len(values) - 1)).astype(int)
            scaled[:, i] = values[idx]
        else:
            scaled[:, i] = base[:, i] * (upper - lower) + lower
    return scaled

    
\end{lstlisting}

\subsection{Normalization and Metrics}
We use feature-normalized $\ell_2$ and $\ell_1$ distances (per-feature scaling by sample std) and LOF-based plausibility filtering/affinity to bias the cost and report affinity/robustness metrics.\\

\lstset{
  basicstyle=\ttfamily\scriptsize,
  breaklines=true,
  columns=fullflexible,
  keepspaces=true,
  showstringspaces=false,
  upquote=true
}
\begin{lstlisting}[language=Python]
def feature_normalized_distance(X, X_prime, std_devs, epsilon=1e-10):
    """
    Feature-normalized Euclidean distance: ||(X - X') / std||_2.
    Returns (n_samples, 1).
    """
    valid = std_devs > epsilon
    if not np.any(valid):
        raise ValueError("All features have near-zero std.")
    diff = (X[:, valid] - X_prime[:, valid]) / std_devs[valid]
    return np.sqrt(np.sum(diff**2, axis=1)).reshape(-1, 1)

def feature_normalized_l1_distance(X, X_prime, std_devs, epsilon=1e-10):
    """
    Feature-normalized L1 distance: ||(X - X') / std||_1.
    Returns (n_samples, 1).
    """
    valid = std_devs > epsilon
    if not np.any(valid):
        raise ValueError("All features have near-zero std.")
    diff = np.abs(X[:, valid] - X_prime[:, valid]) / std_devs[valid]
    return np.sum(diff, axis=1).reshape(-1, 1)

def filter_outliers(new_point, X, n_neighbors=20):
    """
    Returns True if new_point is predicted as inlier by LOF.
    """
    lof = LocalOutlierFactor(n_neighbors=n_neighbors, novelty=True)
    lof.fit(X)
    return lof.predict(new_point.reshape(1, -1)) == 1

def compute_lof_affinity(new_point, X, n_neighbors=20):
    """
    LOF-based affinity in (0, +inf): exp(1 + score); ~1 near inlier threshold.
    """
    lof = LocalOutlierFactor(n_neighbors=n_neighbors, novelty=True)
    lof.fit(X)
    score = lof.score_samples(new_point.reshape(1, -1))[0]
    return np.exp(1 + score)
\end{lstlisting}

%
\end{document}